\documentclass[conference]{IEEEtran}
\pdfoutput=1
\IEEEoverridecommandlockouts
\usepackage{cite}
\usepackage{amsmath,amssymb,amsfonts}
\usepackage{algorithmic}
\usepackage{graphicx}
\usepackage{textcomp}
\usepackage{booktabs}
\usepackage[colorlinks,linkcolor=blue]{hyperref}
\usepackage{multirow}
\usepackage[caption=false]{subfig}
\usepackage{xcolor}
\def\BibTeX{{\rm B\kern-.05em{\sc i\kern-.025em b}\kern-.08em
    T\kern-.1667em\lower.7ex\hbox{E}\kern-.125emX}}
\begin{document}

\title{SACANet: scene-aware class attention network for semantic segmentation of remote sensing images
}

\author{
\IEEEauthorblockN{\textit{Xiaowen Ma}$^{1}$,
\textit{Rui Che}$^{1}$, 
\textit{Tingfeng Hong}$^{1}$,
\textit{Mengting Ma}$^{1}$,
\textit{Ziyan Zhao}$^{1}$,
\textit{Tian Feng}$^{1,2}$\IEEEauthorrefmark{1}\thanks{\IEEEauthorrefmark{1}~Corresponding author (Email: t.feng@zju.edu.cn).\newline This work was supported in part by the National Natural Science Foundation of China under Grant No. 62202421; in part by Zhejiang Provincial Key Research and Development Program under Grant No. 2021C01031; in part by Zhejiang Provincial Natural Science Foundation of China under Grant No. LTGS23F020001; in part by Ningbo Yongjiang Talent Introduction Programme under Grant No. 2021A-157-G; and in part by the Public Welfare Science and Technology Plan of Ningbo City under Grant No. 2022S125.} and
\textit{Wei Zhang}$^{1}$
}

\IEEEauthorblockA{$^{1}$Zhejiang University \quad $^{2}$Alibaba-Zhejiang University Joint Research Institute of Frontier Technologies}

}

\maketitle

\begin{abstract}
Spatial attention mechanism has been widely used in semantic segmentation of remote sensing images given its capability to model long-range dependencies. Many methods adopting spatial attention mechanism aggregate contextual information using direct relationships between pixels within an image, while ignoring the scene awareness of pixels (i.e., being aware of the global context of the scene where the pixels are located and perceiving their relative positions). Given the observation that scene awareness benefits context modeling with spatial correlations of ground objects, we design a scene-aware attention module based on a refined spatial attention mechanism embedding scene awareness. Besides, we present a local-global class attention mechanism to address the problem that general attention mechanism introduces excessive background noises while hardly considering the large intra-class variance in remote sensing images. In this paper, we integrate both scene-aware and class attentions to propose a scene-aware class attention network (SACANet) for semantic segmentation of remote sensing images. Experimental results on three datasets show that SACANet outperforms other state-of-the-art methods and validate its effectiveness. Code is available at~\href{https://github.com/xwmaxwma/rssegmentation}{https://github.com/xwmaxwma/rssegmentation}.
\end{abstract}


\section{Introduction}
Semantic segmentation aims to predict the semantic class (or label) of each pixel, and is one of the fundamental and challenging tasks in remote sensing image analysis. By providing cues on semantic and localization information for the ground objects of interest, semantic segmentation has been regarded as a vital role in the areas of road extraction~\cite{road}, urban planning~\cite{urban}, environmental detection~\cite{environment}, etc. In recent years, convolutional neural networks (CNNs) have facilitated the development of semantic segmentation because of their strength in feature extraction. However, it is limited by the local receptive fields and short-range contextual information due to the fixed geometric structure. Consequently, context modeling, including spatial context modeling and relational context modeling, becomes a noticeable option to capture long-range dependencies. 

Spatial context modeling methods, such as PSPNet~\cite{pspnet} and DeepLabv3+~\cite{deeplabv3+}, integrate spatial context information using spatial pyramid pooling and atrous convolution, respectively. These methods focus on capturing homogeneous context dependencies while often ignoring categorical differences, probably resulting in unreliable contexts if confusing categories are in the scene.

Relational context modeling methods adopt the attention mechanism~\cite{danet, ocrnet, isnet, logcan}, which calculates pixel-level similarity in an image for weighted aggregation of heterogeneous contextual information, and have achieved remarkable results in semantic segmentation tasks. However, these methods concentrate on the relationships between pixels while disregarding their awareness of the scene (i.e., global contextual information and position prior), leaving the spatial correlations of ground objects underexplored in remote sensing images.

In this paper, we firstly refine the spatial attention mechanism and propose a scene-aware attention (SAA) module, which contributes effectively to semantic segmentation of remote sensing images. Besides, remote sensing images are characterized by complex backgrounds and large intra-class variance, whereas the conventional attention mechanism over-introduces the background noises due to dense affinity operations and can hardly deal with intra-class variance. In this regard, we introduce the local-global class attention, which associates pixels with the global class representations using the local class representations as intermediate aware elements, achieving efficient and accurate class-level context modeling.

\begin{figure*}[t]
	\centering
	\includegraphics[width=0.95\textwidth]{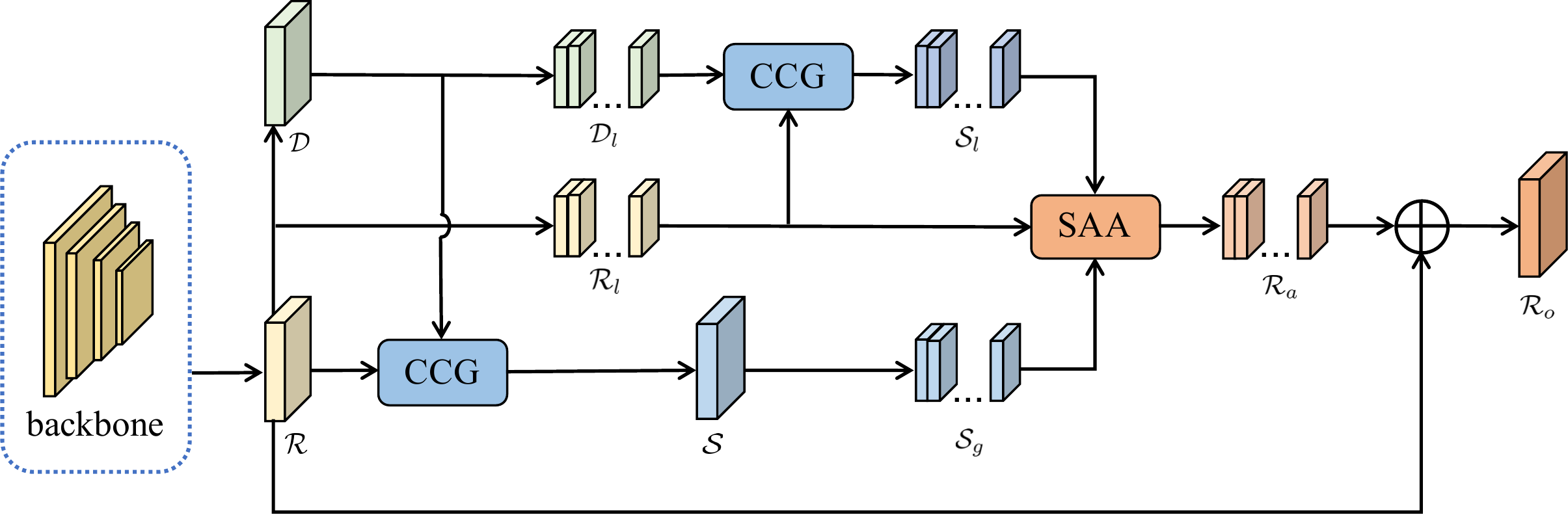}
	\caption{Architecture of the SACANet. Given the extracted feature representations $\mathcal{R}$ from the backbone and the pre-classified representations $\mathcal{D}$, $\mathcal{R}_l$ and $\mathcal{D}_l$ are obtained after spatial dimensional splitting. Global class center $\mathcal{S}_g$ and local class center $\mathcal{S}_l$ are generated by the class center generation (CCG) module. Then, $\mathcal{R}_l$, $\mathcal{S}_l$, and $\mathcal{S}_g$ are input to the SAA module to obtain the enhanced representations $\mathcal{R}_a$. $\mathcal{R}_a$ is recovered to its original spatial dimension and concatenated with $\mathcal{R}$ to obtain the output representations $\mathcal{R}_o$.}
	\label{fig:whole}
\end{figure*}

Our contributions are as follows.
\begin{itemize}
\item We improve the attention mechanism by embedding the scene awareness of pixels to exploit the spatial correlations of ground objects, which is for the first time to integrate the attention mechanism and scene awareness into a unified module for semantic segmentation of remote sensing images.

\item We introduce the local-global class attention mechanism associating pixels with global class representations using local class representations, which achieves efficient and accurate class-level context modeling and tackles complex backgrounds and large intra-class variance in remote sensing images.

\item We propose a scene-aware class attention network (SACANet) combining the class attention with scene awareness for semantic segmentation of remote sensing images. Experimental results show that SACANet achieves the state-of-the-art performance on three benchmark datasets, while reaching a decent trade-off between efficiency and accuracy.
\end{itemize}

\section{Method}

The proposed SACANet, whose architecture is depicted by Fig.~\ref{fig:whole}, comprises three major components: feature extraction, class center generation and scene-aware attention. To begin with, the HRNetv2-w32~\cite{hrnet} pretrained on ImageNet is deployed as the backbone to extract the feature representations $\mathcal{R}$ from an input image, followed by $\mathcal{R}$ being pre-classified to obtain $\mathcal{D}$. The class center generation (CCG) module then takes as input both $\mathcal{R}$ and $\mathcal{D}$ to achieve the global class center $\mathcal{S}$, which is further cropped to output $\mathcal{S}_g$. Likewise, $\mathcal{R}_l$ and $\mathcal{D}_l$, which are obtained by cropping $\mathcal{R}$ and $\mathcal{D}$, are processed by the CCG module to achieve the local class center $\mathcal{S}_l$. Furthermore, the scene-aware attention (SSA) module takes $\mathcal{R}_l$, $\mathcal{S}_l$, $\mathcal{S}_g$ as inputs to obtain the enhanced feature representations $\mathcal{R}_a$. After original spatial dimensions are recovered, $\mathcal{R}_a$ and $\mathcal{R}$ are concatenated to achieve the output feature representations $\mathcal{R}_o$. The final segmentation map is output after quadruple upsampling.

\begin{figure*}[t]
	\centering
	\includegraphics[width=0.9\textwidth]{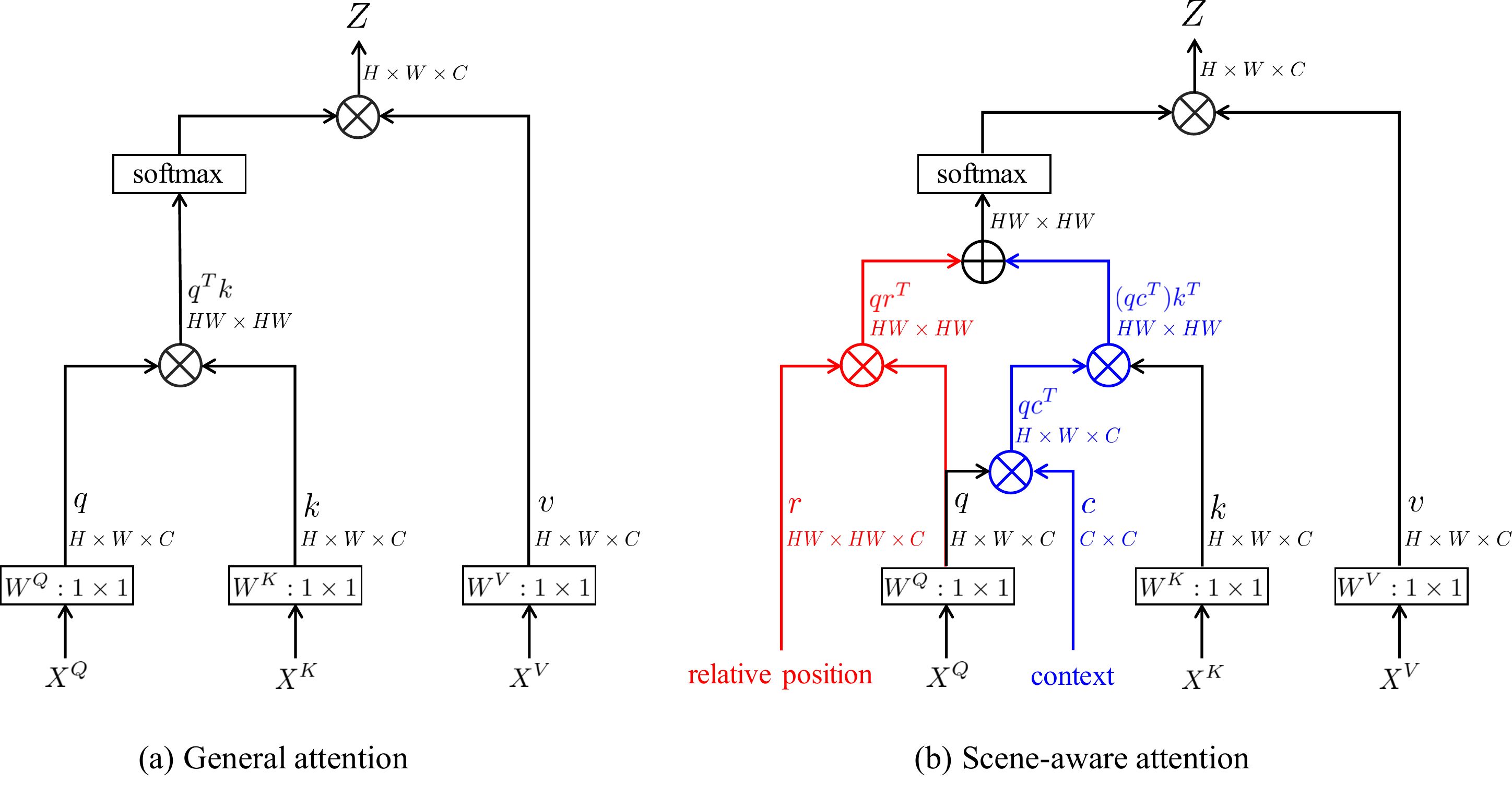}
	\caption{Illustration of (a) the general attention (GA) and (b) the scene-aware attention (SAA). The red and blue parts in the SAA are newly added compared to the GA, representing the relative position and context to embed the pixels' scene awareness in the attention.}
	\label{fig:saa}
\end{figure*}

To illustrate our design of scene-aware attention and local-global class attention, we describe the general form of the attention mechanism as follows: Given feature representations $X^Q, X^K, X^V \in \mathbb{R} ^{H\times W\times \hat{C}}$, where $H$, $W$ and $\hat{C}$ denote height, width and dimension of the feature representations respectively. As shown in Fig.~\ref{fig:saa}, the attention mechanism applies three different $1\times1$ convolutions $W^Q, W^K, W^V \in \mathbb{R} ^{\hat{C}\times C}$ to obtain $q, k, v \in \mathbb{R} ^{H\times W\times C}$ as follows,
\begin{equation}
    q=X^QW^Q,   k=X^KW^K,    v=X^VW^V.
\end{equation}

Each output element $\emph{Z}_i$ is a weighted sum of input elements $\{v_j\}$ as follows,
\begin{equation}
    \emph{Z}_{i}=\sum_{j=1}^{H\times W}\alpha_{ij}v_j,
\end{equation}
where $\alpha_{ij}$ denotes the weight from the softmax function on $e_{ij}$, which is obtained by a scaled dot-product attention as follows,
\begin{equation}
   e_{ij}=\frac{q_i{k_j}^T}{\sqrt{C}}.
\end{equation}

\subsection{Scene-Aware Attention}
Ground objects in remote sensing images have intrinsic spatial correlations that are frequently observable. For example, vehicles are usually found to stay on a road; buildings are densely distributed on both sides of a road. Therefore, it is supposed to facilitate the modeling of corresponding patterns by embedding the scene awareness of pixels (i.e., considering the global context of pixels as well as their relative position in the attention).

\textbf{Contextual Information Embedding.} In remote sensing images, pair-wise relationships between ground objects may vary in different scenes. For example, a road that usually coexists with buildings in an urban area may be surrounded by cropland in a rural area. It suggests that embedding contextual information can benefit the modeling of pixel-level relationships. Inspired by~\cite{cbam}, we propose a context matrix and reformulate the previous equation as follows,
\begin{equation}
   e_{ij}=\frac{(q_ic){k_j}^T}{\sqrt{C}},
\end{equation}
and the context matrix $c$ is computed as follows,
\begin{align}
   c=diag(\sigma(W_1(&W_0(AvgPool(Q)))\nonumber \\
   &+W_1(W_0(MaxPool(Q)))))
\end{align}
where $\sigma$ denotes the sigmoid function, $W_0 \in \mathbb{R} ^{(C/\epsilon)\times C}$, $W_1 \in \mathbb{R} ^{C \times (C/\epsilon)}$, and $\epsilon$ is the reduction ratio. We create a diagonal matrix for the vector to connect the summarized contextual information with the input features. Finally, the context matrix $c$ contextualizes the input features $q$ so that the attention can be adjusted by the given context.

\textbf{Position Prior Embedding.} In remote sensing images, ground objects are spatially distributed following specific intrinsic patterns. In particular, certain combinations or concurrences usually occur to the objects in close proximity, and the pixels in an object's vicinity may demonstrate high correlation. These observations suggest that a pixel's awareness of the scene relies on its sensitivity to relative positions, which are embedded as,
\begin{equation}
   e_{ij}=\frac{(q_ic){k_j}^T+q_i r_{ij}^T}{\sqrt{C}}.
\end{equation}
Unlike previous work in the field of natural language processing~\cite{relative}, we extend the words in a one-dimensional sequence to the pixels in a two-dimensional plane, considering their relative positions as a combined effect along both horizontal and vertical directions. Specifically, the encoding of relative position $r_{ij}$ is defined as follows,
\begin{equation}
   r_{ij}=P_{I^x(i,j), I^y(i,j)},
\end{equation}
where $P \in \mathbb{R} ^{(2\xi+1)\times (2\xi+1)\times C}$ is a bucket storing a set of indexed trainable vectors, $I^x(i,j)=g(x_i-x_j)$ and $I^y(i,j)=g(y_i-y_j)$ represent subscripts for horizontal and vertical directions, forming two-dimensional indices for $P$, and $g$ denotes an index function as follows,
\begin{equation}
   g(x)=max(-\xi, min(x,\xi)),
\end{equation}
where $\xi$ refers to the maximum pixel-level distance. Actually, $g(x)$ maps the distance to an integer in finite set, largely reducing the number of parameters and computation cost needed for high resolution remote sensing images.

\begin{figure}[t]
	\centering
	\includegraphics[width=0.45\textwidth]{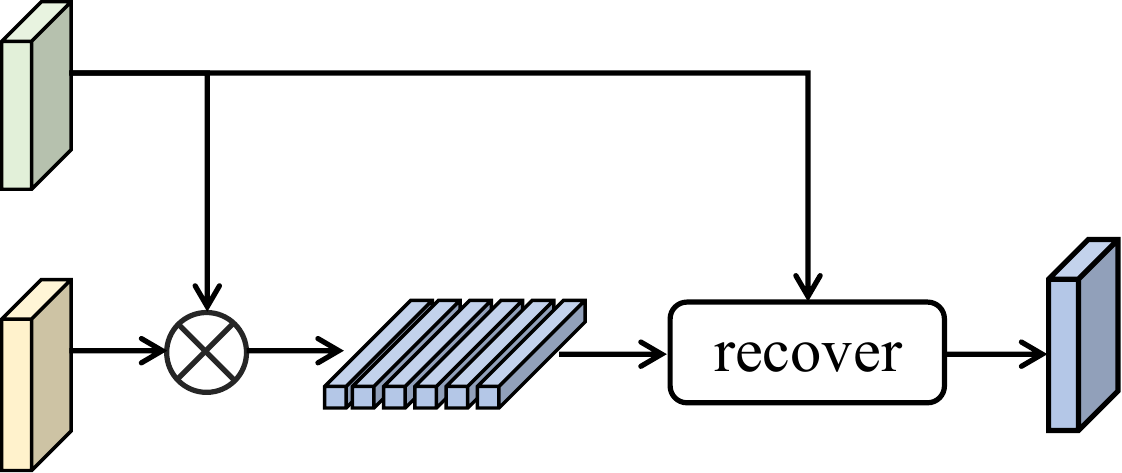}
	\caption{Architecture of the CCG.}
	\label{fig:CCG}
\end{figure}

% &Background &Building &Road &Water &Barren &Forest &Agriculture
\setlength{\tabcolsep}{5pt}
\begin{table*}[t]
	\begin{center}
		\caption{
		Comparison with the state-of-the-art methods on the test set from the LoveDA, ISPRS Vaihingen and ISPRS Potsdam datasets. Per-class best performance is marked in bold.
		}
		\label{table:1}
		\begin{tabular}{c|cccccccc|ccc|ccc}
		\toprule
		    \multirow{2}{*}{Method} &\multicolumn{8}{c|}{LoveDA} &\multicolumn{3}{c|}{Vaihingen} &\multicolumn{3}{c}{Potsdam}\\
			&Back. &Buil. &Road &Water &Barren &Forest &Agri. &mIoU  &AF &mIoU &OA &AF &mIoU &OA\\
			\midrule
			PSPNet~\cite{pspnet}  & 44.4 & 52.1 &53.5 &76.5 &9.7 &44.1 &57.9 &48.3 &86.47 &76.78 &89.36 & 89.98 &81.99 &90.14\\
			DeepLabv3+~\cite{deeplabv3+} & 43.0 & 50.9 &52.0 &74.4 &10.4 &44.2 &58.5 &47.6 &86.77 &77.13 &89.12 &90.86 &84.24 &89.18\\
			DANet~\cite{danet} &44.8 &55.5 &53.0 &75.5 &17.6 &45.1 &60.1 &50.2 &86.88 &77.32 &89.47 &89.60 &81.40 &89.73\\
			Semantic FPN~\cite{semanticfpn} &42.9 & 51.5 &53.4 &74.7 &11.2 &44.6 &58.7 &48.2 &87.58 &77.94 &89.86 & 91.53 &84.57 &90.16\\
			FarSeg~\cite{farseg} &43.1 &51.5 &53.9 &76.6 &9.8 &43.3 &58.9 &48.2 &87.88 &79.14 &89.57  &91.21 & 84.36 &89.87\\
			OCRNet~\cite{ocrnet} &44.2 &55.1 &53.5 &74.3 &18.5 &43.0 &60.5 &49.9 &89.22 &81.71 &90.47 &92.25 &86.14 &90.03\\
			LANet~\cite{lanet}  &40.0 &50.6 &51.1 &78.0 &13.0 &43.2 &56.9 &47.6 &88.09 &79.28 &89.83 &91.95 &85.15 &90.84\\
			ISNet~\cite{isnet}&44.4 &57.4 &58.0 &77.5 &\bf21.8 &43.9 &60.6 &51.9 &90.19 &82.36 &90.52 &92.67 &86.58 &91.27 \\
			Segmenter~\cite{segmenter} &38.0 &50.7 &48.7 &77.4 &13.3 &43.5 &58.2 &47.1 &88.23 &79.44 &89.93 &92.27 &86.48 &91.04\\
			SwinUperNet~\cite{swintransformer} &43.3 &54.3 &54.3 &78.7 &14.9 &45.3 &59.6 &50.0 &89.9 &81.8 &91.0 &92.24 &86.37 &90.98\\
			MANet~\cite{manet} &38.7 &51.7 &42.6 &72.0 &15.3 &42.1 &57.7 &45.7 &90.41 &82.71 &90.96 &92.90 &86.95 &91.32\\
			FLANet~\cite{flanet} & 44.6 & 51.8 &53.0 &74.1 &15.8 &45.8 &57.6 &49.0 &87.44 &78.08 &89.60 &93.12 &87.50 &91.87\\
		      ConvNeXt~\cite{convnext}  & 46.9 & 53.5 &56.8 &76.1 &15.9 &\bf47.5 &61.8 &51.2 &90.50 &82.87 &91.36 & 93.03 &87.17 &91.66\\
			PoolFormer~\cite{poolformer} & 45.8 & 57.1 &53.3 &\bf80.7 &19.8 &45.6 &64.5 &52.4 &89.59 &81.35 &90.30 &92.62 &86.45 &91.12\\
			\midrule
			SACANet (Ours) &\bf47.6 &\bf59.1 &\bf58.4 &80.5 &17.8 &46.7 &\bf67.1 &\bf53.9 &\bf91.68 &\bf84.49 &\bf92.10 &\bf93.64 &\bf87.89 &\bf92.28\\
			\bottomrule
		\end{tabular}
	\end{center}
\end{table*}
\setlength{\tabcolsep}{2pt}

\subsection{Local-Global Class Attention}
Self-attention mechanism~\cite{danet,swintransformer,nonlocal} has so far dominated the attention-based semantic segmentation methods, whose inputs $X^Q$, $X^K$ and $X^V$ to the attention module are all set to the given feature representations $\mathcal{R} \in \mathbb{R} ^{H\times W\times \hat{C}}$.

Considering that remote sensing images are characterized by complex backgrounds and large intra-class variance, these methods can result in massive background noise towards poor performances due to dense affinity operations. Several class attention methods attempt to resolve this problem using global class center for class-level context modeling; However, they are yet to consider intra-class variance and the case that pixels may be semantically distant from the global class center impacting on the class context modeling. Therefore, we present the local-global class attention to improve the performance of class-level context modeling. In particular, pixels are indirectly associated with global class representations by introducing local class representations.

As shown in Fig.~\ref{fig:CCG}, for the feature representations $\mathcal{R}\in \mathbb{R} ^{\hat{C}\times H\times W}$, a pre-classification is deployed (i.e., two consecutive $1\times 1$
convolution layers) to obtain the corresponding distribution $\mathcal{D}\in \mathbb{R} ^{K\times H\times W}$, where $K$ is the number of classes. The global class center $S$
is defined as follows,
\begin{equation}
\mathcal{S}= \psi(\mathcal{D}^{K\times (H\times W)}\otimes \mathcal{R}^{(H\times W)\times \hat{C}}) ,
\end{equation} 
where $\mathcal{S}$ denotes a $H\times W\times \hat{C}$ matrix, $\psi$ represents a function to place class centers in the original feature map according to the pre-classification generated mask. Then, $\mathcal{R}$ and $\mathcal{D}$ are split along the spatial dimension to reach $\mathcal{R}_l$ and $\mathcal{D}_l$, followed by calculating the local class representations $\mathcal{S}_l$ as follows,
\begin{equation}
\mathcal{S} _{l} = \psi(\mathcal{D} _{l}^{(N_h\times N_w) \times K\times ( h \times w)}\otimes \mathcal{R} _{l}^{(N_h\times N_w) \times ( h \times w)\times C}) ,
\end{equation} 
where $h$ and $w$ are the height and width of the selected local patch, $N_h = \frac{H}{h}$, and $N_w = \frac{W}{w}$. Similarly, $\mathcal{S}$ is split along the spatial dimension to obtain $\mathcal{S}_g\in \mathbb{R} ^{(N_h\times N_w) \times (h\times w)\times \hat{C}}$. Hence, the inputs to the attention module are as follows:
\begin{equation}
   X^Q=\mathcal{R}_l,\quad X^K=\mathcal{S}_l,\quad X^V=\mathcal{S}_g.
\end{equation}

The design of local-global class attention combines scene awareness with local-global class attention. In addition, the slicing operation provides a noticeable decrease in the number of parameters and computation, enabling the lightweight of the method.

\section{EXPERIMENTAL RESULTS}
\subsection{Datasets and Evaluation Metrics}
We conduct the experiments on three publicly available datasets to evaluate our SACANet in three common metrics: average F1-score (AF), mean Intersection-over-Union (mIoU), and overall accuracy (OA).
\setlength{\tabcolsep}{4pt}
\begin{table}[t]
	\begin{center}
		\caption{Comparison with other popular context aggregation modules.
		}
		\label{table:2}
		\begin{tabular}{c|ccc}
		\toprule
			Method &Params (M) &FLOPs (G) &Memory (MB)\\
			\midrule
			PPM~\cite{pspnet} & 23.1 & 309.5 & 257 \\
			ASPP~\cite{deeplabv3+} & 15.1 & 503.0 & 284 \\
			DAB~\cite{danet} &23.9 & 392.2 & 1546\\
			OCR~\cite{ocrnet} &10.5 & 354.0 &202 \\
			PAM+AEM~\cite{lanet} &10.4 & 157.6 &489\\
			ILCM+SLCM~\cite{isnet} &11.0 &180.6 &638\\
   FLA~\cite{flanet} &11.5 &154.9 &645\\
			
			\midrule
			CCG+SAA (Ours) &\bf2.7 &\bf44.4 &\bf76\\
			\bottomrule
		\end{tabular}
	\end{center}
\end{table}
\vspace{5mm}
\setlength{\tabcolsep}{2pt}

LoveDA~\cite{wang2021loveda} contains 5987 fine-resolution optical remote sensing images (GSD 0.3 m) at a size of 1024 × 1024 pixels and includes 7 landcover categories, i.e., building, road, water, barren, forest, agriculture and background. Specifically, we use 2522 images for training, 1669 images for validation and the remaining 1796 images for testing.

ISPRS Vaihingen~\cite{www.isprs.org} contains 33 TOP tiles and DSMs (GSD 9 cm) collected from a small village and includes 6 landcover categories, i.e., impervious surfaces, building, low vegetation, tree, car, and clutter/background. The size of the images varies from 1996 × 1995 pixels to 3816 × 2550 pixels. We use 16 images for training and the remaining 17 for testing. 

ISPRS Potsdam~\cite{www.isprs.org} consists of 38 TOP tiles and DSMs (GSD 5 cm) collected from a historic city at a size of 6000 × 6000 pixels and includes the same six categories as the Vaihingen dataset. We use 24 images for training and the remaining 14 for testing. 

\subsection{Implementation Details}
For all experiments, the optimizer is SGD with the batch size of 4, and the initial learning rate is set to 0.01 with a poly decay strategy and a weight decay of 0.0001. Following previous work~\cite{lanet,manet}, we randomly crop the images from three datasets to produce $512\times 512$ patches, and the augmentation methods, such as random scale ([0.5, 0.75, 1.0, 1.25, 1.5]), random vertical flip, random horizontal flip and random rotate, are adopted in the training process. The number of epochs on LoveDA, ISPRS Vaihingen, and ISPRS Potsdam is set to 30, 150 and 80, respectively.

\begin{figure*}[t]
\centering
\captionsetup[subfloat]{labelsep=none,format=plain,labelformat=empty}
\subfloat[imgae]{
\begin{minipage}[t]{0.14\linewidth}
\includegraphics[width=1\linewidth]{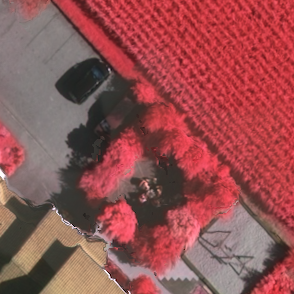}\vspace{1pt}
\includegraphics[width=1\linewidth]{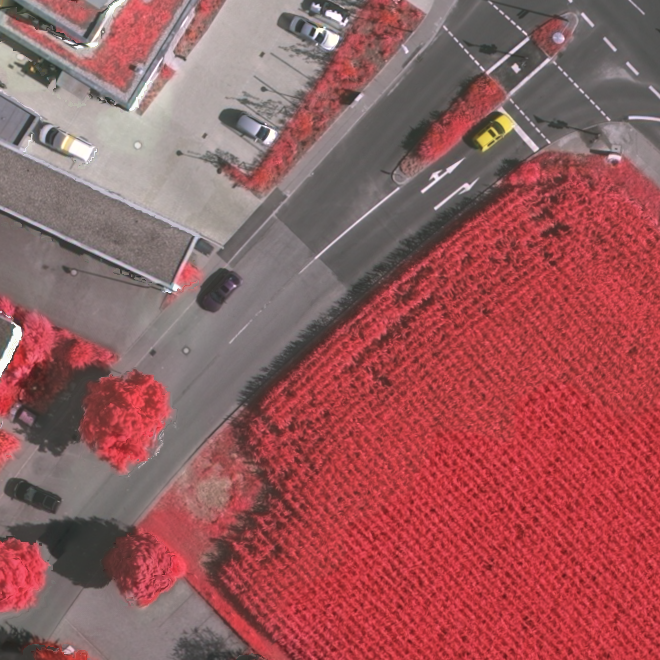}\vspace{3pt}
\includegraphics[width=1\linewidth]{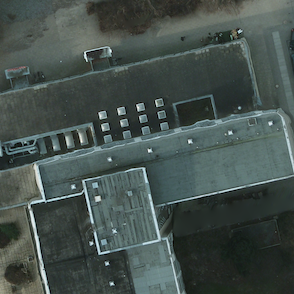}\vspace{1pt}
\includegraphics[width=1\linewidth]{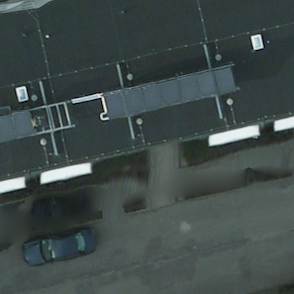}\vspace{1pt}
\end{minipage}}
\subfloat[GT]{
\begin{minipage}[t]{0.14\linewidth}
\includegraphics[width=1\linewidth]{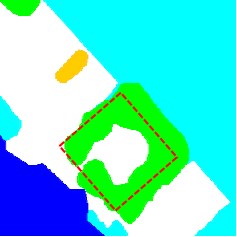}\vspace{1pt}
\includegraphics[width=1\linewidth]{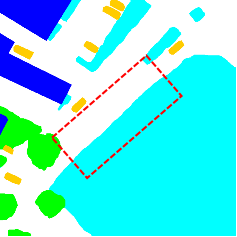}\vspace{3pt}
\includegraphics[width=1\linewidth]{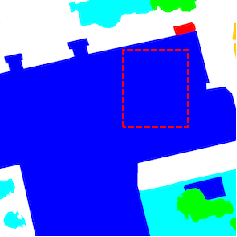}\vspace{1pt}
\includegraphics[width=1\linewidth]{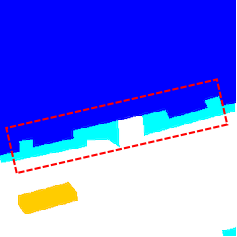}\vspace{1pt}
\end{minipage}}
\subfloat[PSPNet]{
\begin{minipage}[t]{0.14\linewidth}
\includegraphics[width=1\linewidth]{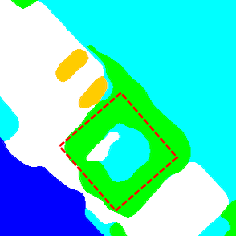}\vspace{1pt}
\includegraphics[width=1\linewidth]{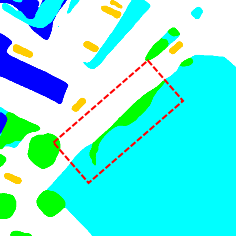}\vspace{3pt}
\includegraphics[width=1\linewidth]{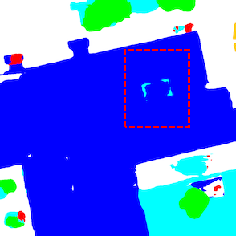}\vspace{1pt}
\includegraphics[width=1\linewidth]{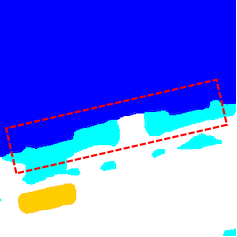}\vspace{1pt}
\end{minipage}}
\subfloat[DANet]{
\begin{minipage}[t]{0.14\linewidth}
\includegraphics[width=1\linewidth]{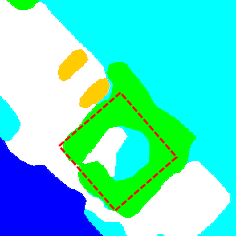}\vspace{1pt}
\includegraphics[width=1\linewidth]{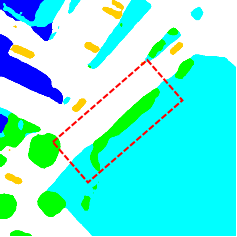}\vspace{3pt}
\includegraphics[width=1\linewidth]{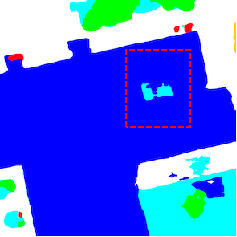}\vspace{1pt}
\includegraphics[width=1\linewidth]{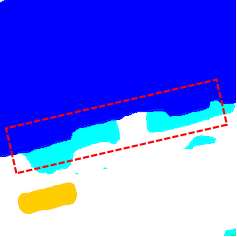}\vspace{1pt}
\end{minipage}}
\subfloat[MANet]{
\begin{minipage}[t]{0.14\linewidth}
\includegraphics[width=1\linewidth]{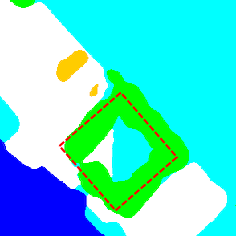}\vspace{1pt}
\includegraphics[width=1\linewidth]{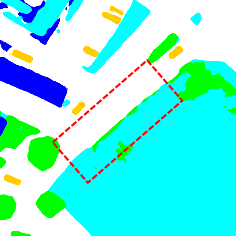}\vspace{3pt}
\includegraphics[width=1\linewidth]{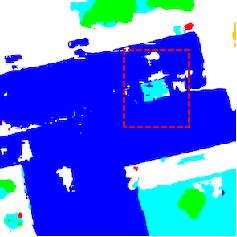}\vspace{1pt}
\includegraphics[width=1\linewidth]{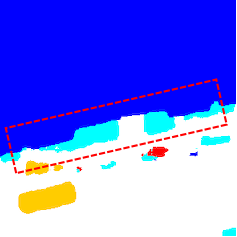}\vspace{1pt}
\end{minipage}}
\subfloat[SACANet]{
\begin{minipage}[t]{0.14\linewidth}
\includegraphics[width=1\linewidth]{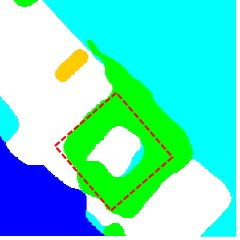}\vspace{1pt}
\includegraphics[width=1\linewidth]{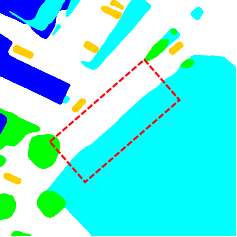}\vspace{3pt}
\includegraphics[width=1\linewidth]{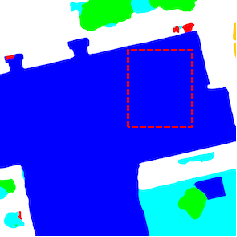}\vspace{1pt}
\includegraphics[width=1\linewidth]{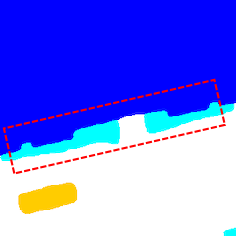}\vspace{1pt}
\end{minipage}}
\caption{Example outputs from the SACANet and other methods on the ISPRS Vaihingen test set and ISPRS Potsdam test set. Best viewed in color and zoom in.}
\label{fig:res}
\end{figure*}
\subsection{Comparison and Analysis}
As shown in Table~\ref{table:1}, the proposed method outperforms other state-of-the-art methods. Specifically, our SACANet achieves an improvement of 1.5\% in mIoU compared to PoolFormer~\cite{poolformer} on LoveDA; On ISPRS Vaihingen and Potsdam, SACANet's improvement is about 3.1\% and 1.4\%, respectively. In particular, significant improvements are found to the background class, which is complex and has a large intra-class variance, as well as the road and agriculture classes, which are closely related to the corresponding scene, on LoveDA. These improvements have validated the effectiveness of SACANet on semantic segmentation of remote sensing images via embedding scene-aware and local-global attentions. Fig.~\ref{fig:res} shows example results from PSPNet, DANet, MANet and our SACANet. The proposed method not only better preserves the integrity and regularity of semantic objects such as building and low vegetation, but also improves the segmentation performance of small objects, such as car.

\setlength{\tabcolsep}{12pt}
\begin{table}[t]
	\begin{center}
		\caption{
		      Ablation study of SACANet on ISPRS Vaihingen test set. Highest scores are marked in bold. All scores are reported in percentage.
		}
		\label{table:3}
		\begin{tabular}{l|cccc}
			\toprule
			Model &AF &mIoU &OA\\
			\midrule
			 Base & 89.65& 81.87&90.13 \\
			 Base+GA &89.34 &81.62 &89.94 \\
			 Base+SAA &90.84 &83.21 &91.57\\
			 Base+CCG(g+g)+GA &89.60 &82.03 &90.13\\
              Base+CCG(l+g)+GA &90.48 &82.76 &91.23\\
              Base+CCG(g+g)+SAA &90.67 &83.47 &91.12\\
			 \midrule
			 SACANet (Ours) &\bf91.68 &\bf84.49 &\bf92.10\\
			\bottomrule
		\end{tabular}
	\end{center}
\end{table}

In addition, we compare several context aggregation modules and ours in three metrics: number of parameters (Params) measured in million (M), number of floating-point operations per second (FLOPs) measured in giga (G), and the memory consumption (Memory) measured in megabytes (MB). As shown in Table~\ref{table:2}, the attention modules in our method are based on local patches, greatly reducing the required number of parameters and computation. Specifically, we require only 26\% of Params, 13\% of the FLOPs, and 38\% of the Memory compared to the light OCR~\cite{ocrnet} module, which significantly improves the efficiency of the model.

Furthermore, we conduct an ablation study using HRNetv2-w32 as the base on ISPRS Vaihingen to investigate the impacts of SAA and CCG modules. As shown in Table~3, we observe that the semantic segmentation performance of the base is slightly degraded after adding the general attention (GA) module, but improved by introducing our SAA module. Besides, our SACANet integrating both SAA and CCG modules achieves an even higher performance. The results of the ablation study have supported that introducing scene-aware and class attentions can benefit semantic segmentation of remote sensing images.

\section{conclusion}
In this paper, we present scene-aware attention that refines the spatial attention mechanism to exploit the inherent spatial correlation of ground objects in remote sensing images. Considering the complex backgrounds and large intra-class variances, we introduce local-global class attention for class-wise context modeling, which prevents dense attention from over-introducing the interference of background noises. Integrating both scene-aware and local-global class attentions, we propose SACANet that significantly improves the semantic segmentation performance on remote sensing images. Experimental results reveal our SACANet's outperformance compared to other state-of-the-art methods. Besides, the proposed method reduces the number of parameters and computation significantly, and achieves a better trade-off between accuracy and efficiency.

\bibliographystyle{IEEEtran}
\bibliography{reference}

\end{document}